\pdfoutput=1

\documentclass{article}

\usepackage[square,sort,comma,numbers]{natbib}

\usepackage[final]{nips_2017}

\usepackage[utf8]{inputenc} 
\usepackage[T1]{fontenc}    
\usepackage{hyperref}       
\usepackage{url}            
\usepackage{booktabs}       
\usepackage{amsfonts}       
\usepackage{nicefrac}       
\usepackage{microtype}      

\usepackage{times}
\usepackage{epsfig}
\usepackage{graphicx}
\usepackage{amsmath}
\usepackage{amssymb}
\usepackage{algorithm}
\usepackage{algorithmic}
\usepackage{subfigure}
\usepackage{cases}
\usepackage{wrapfig}
\usepackage{color}
\usepackage[english]{babel}
\graphicspath{{images/}}

\title{Feature Incay for Representation Regularization}

\author{Yuhui Yuan$^{\dag}$ \hspace{6mm} Kuiyuan Yang$^{\ddag}$  \hspace{6mm}   Chao Zhang$^{\dag}$\\
$^{\dag}$Key Laboratory of Machine Perception(MOE),Peking University\\
$^{\ddag}$Microsoft Research\\
{\tt\small yhyuan@pku.edu.cn, kuyang@microsoft.com, chzhang@cis.pku.edu.cn}
}

\begin{document}

\maketitle
\begin{abstract}
Softmax loss is widely used in deep neural networks for multi-class classification, where each class is represented by a weight vector, a sample is represented as a feature vector, and the feature vector has the largest projection on the weight vector of the correct category when the model correctly classifies a sample. To ensure generalization, weight decay that shrinks the weight norm is often used as regularizer. Different from traditional learning algorithms where  features are fixed and only weights are tunable, features are also tunable as representation learning in deep learning. Thus, we propose feature incay to also regularize representation learning, which favors feature vectors with large norm when the samples can be correctly classified. With the feature incay, feature vectors are further pushed away from the origin along the direction of their corresponding weight vectors, which achieves better inter-class separability. In addition, the proposed feature incay encourages intra-class compactness along the directions of weight vectors by increasing the small feature norm faster than the large ones. Empirical results on MNIST, CIFAR10 and CIFAR100 demonstrate feature incay can improve the generalization ability.
\end{abstract}

\section{Introduction}

\begin{figure*}

\footnotesize
\subfigure[]{
  \label{fig:fig1:softmax_iter200}
  {\includegraphics[width=4.0cm,bb=28 13 537 405]{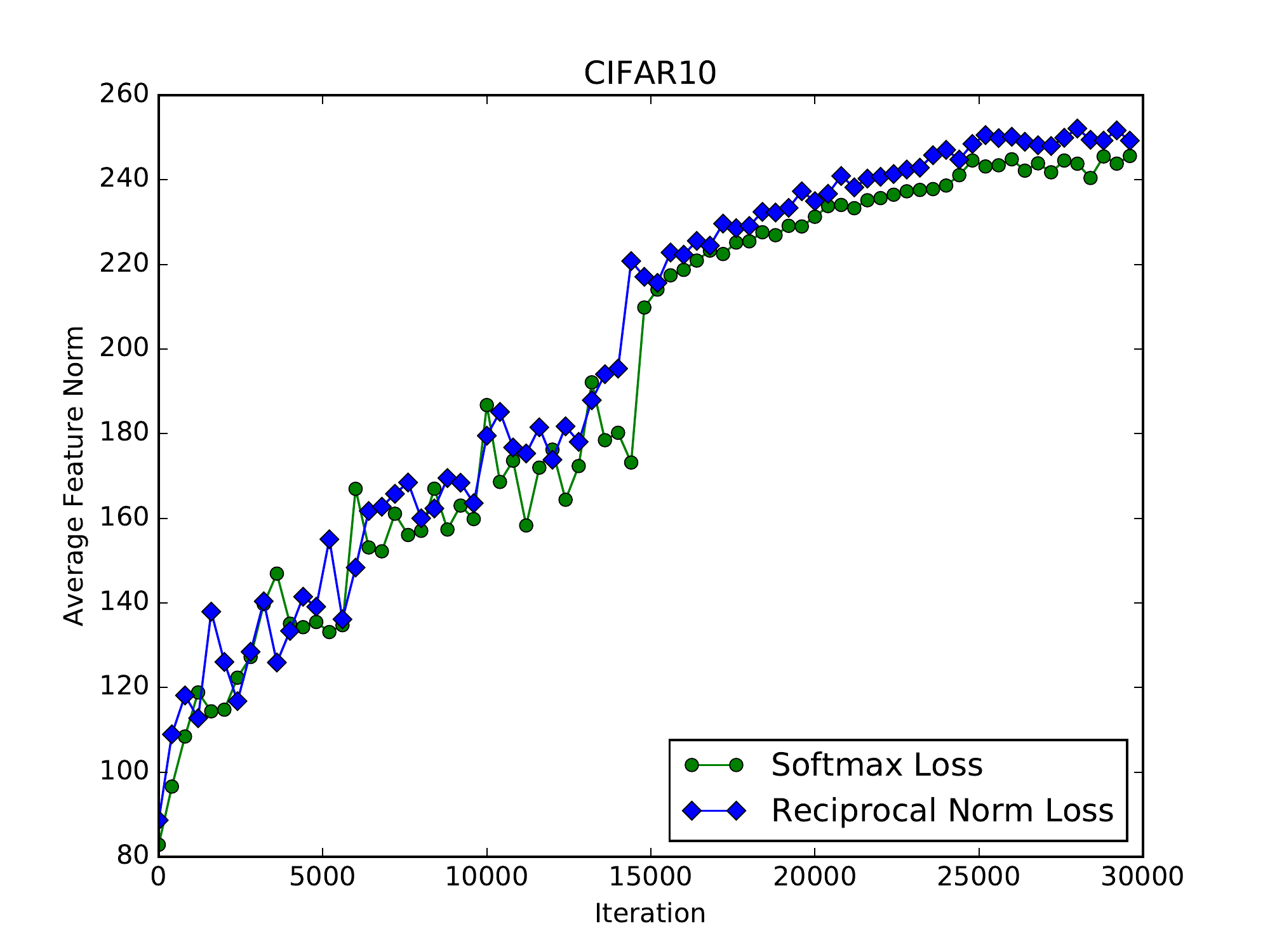}}
  \hspace{5mm}
}
\subfigure[]{
  \label{fig:fig1:softmax_iter2000}
  {\includegraphics[width=4.0cm,bb=25 13 537 405]{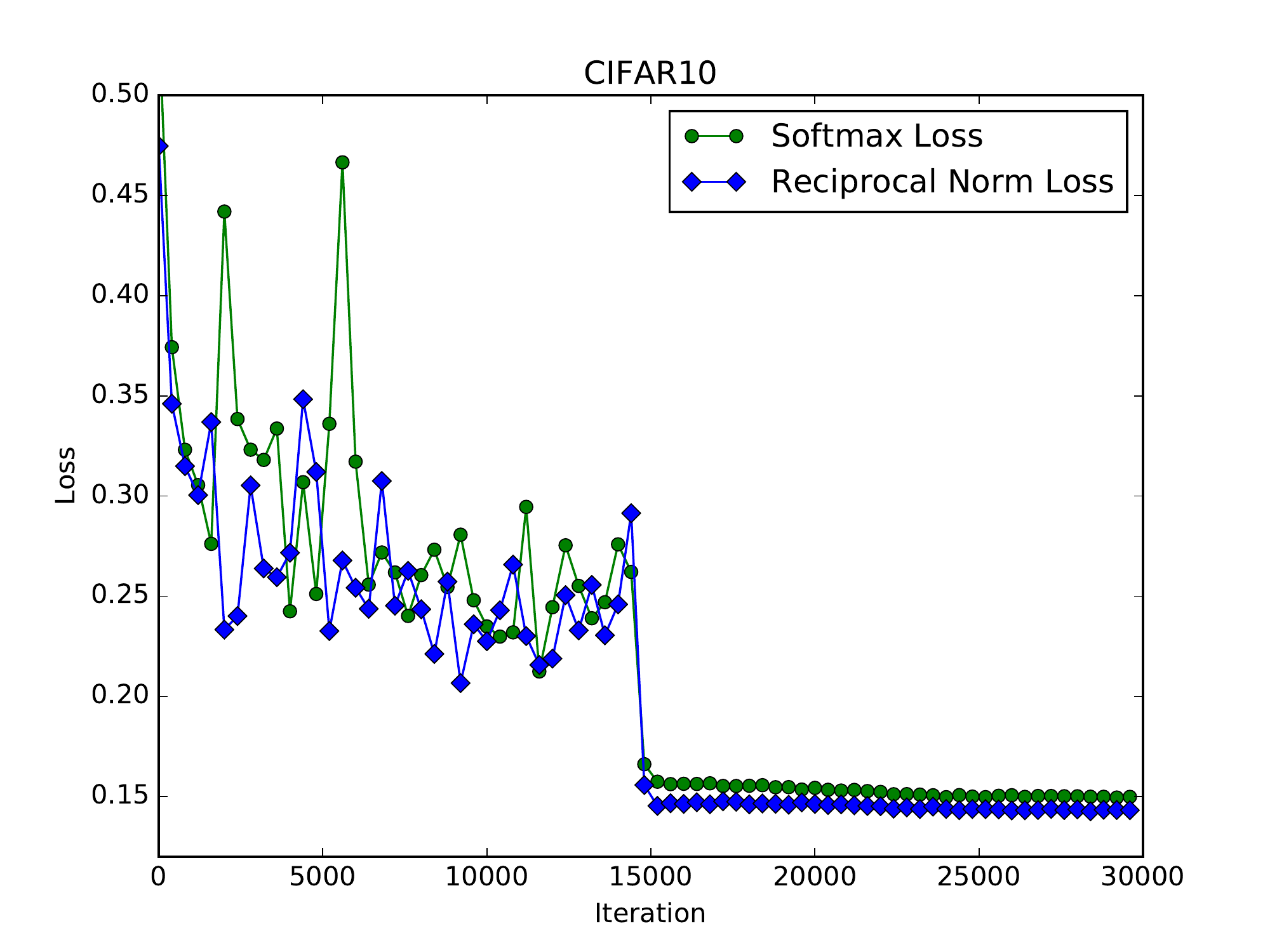}}
  \hspace{5mm}
}
\subfigure[]{
  \label{fig:fig1:softmax_iter20000}
  {\includegraphics[width=4.0cm,bb=25 13 537 405]{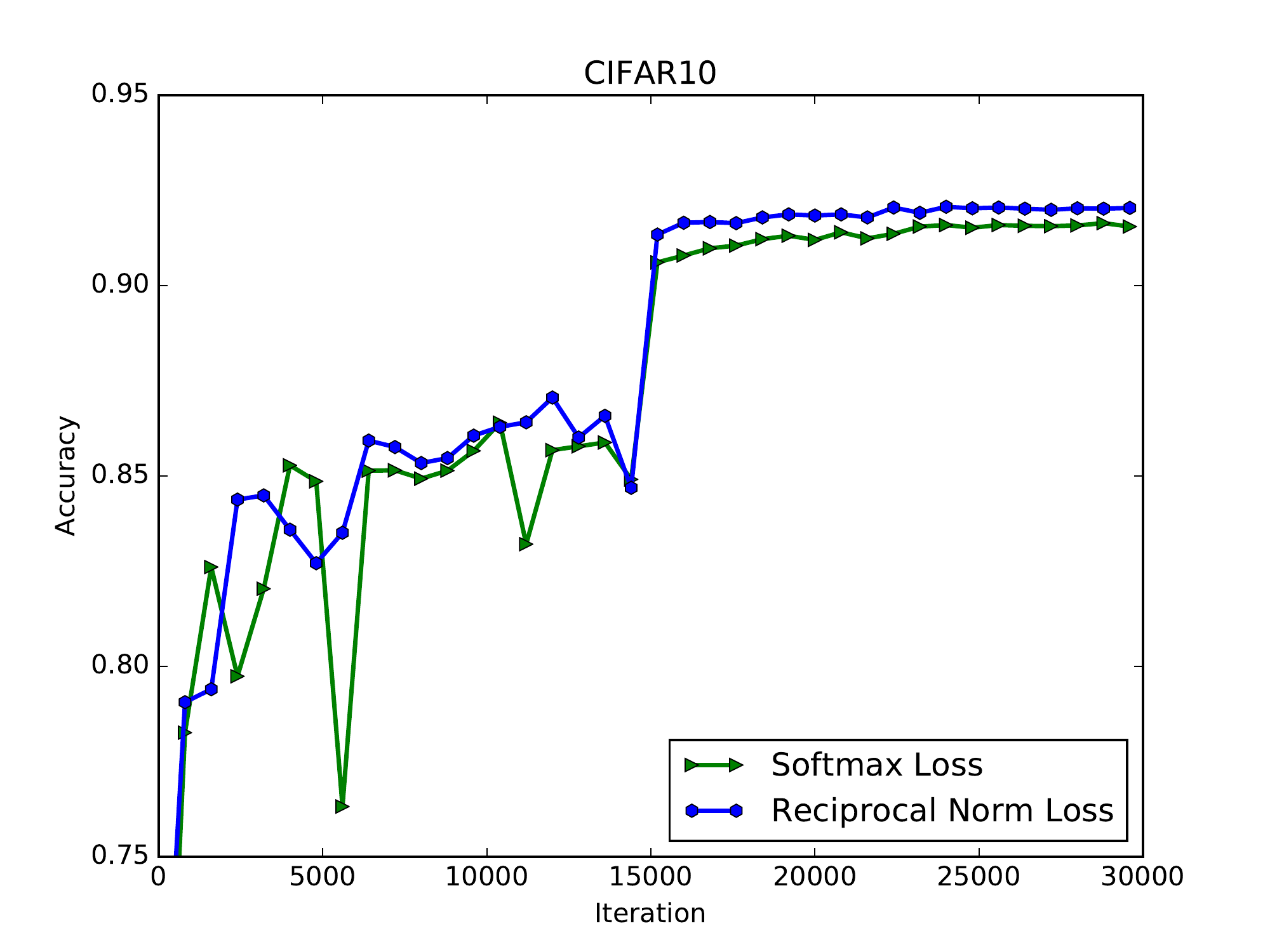}}
}
\subfigure[]{
  \label{fig:fig1:margin1}
  {\includegraphics[width=4.0cm,bb=9 11 411 275]{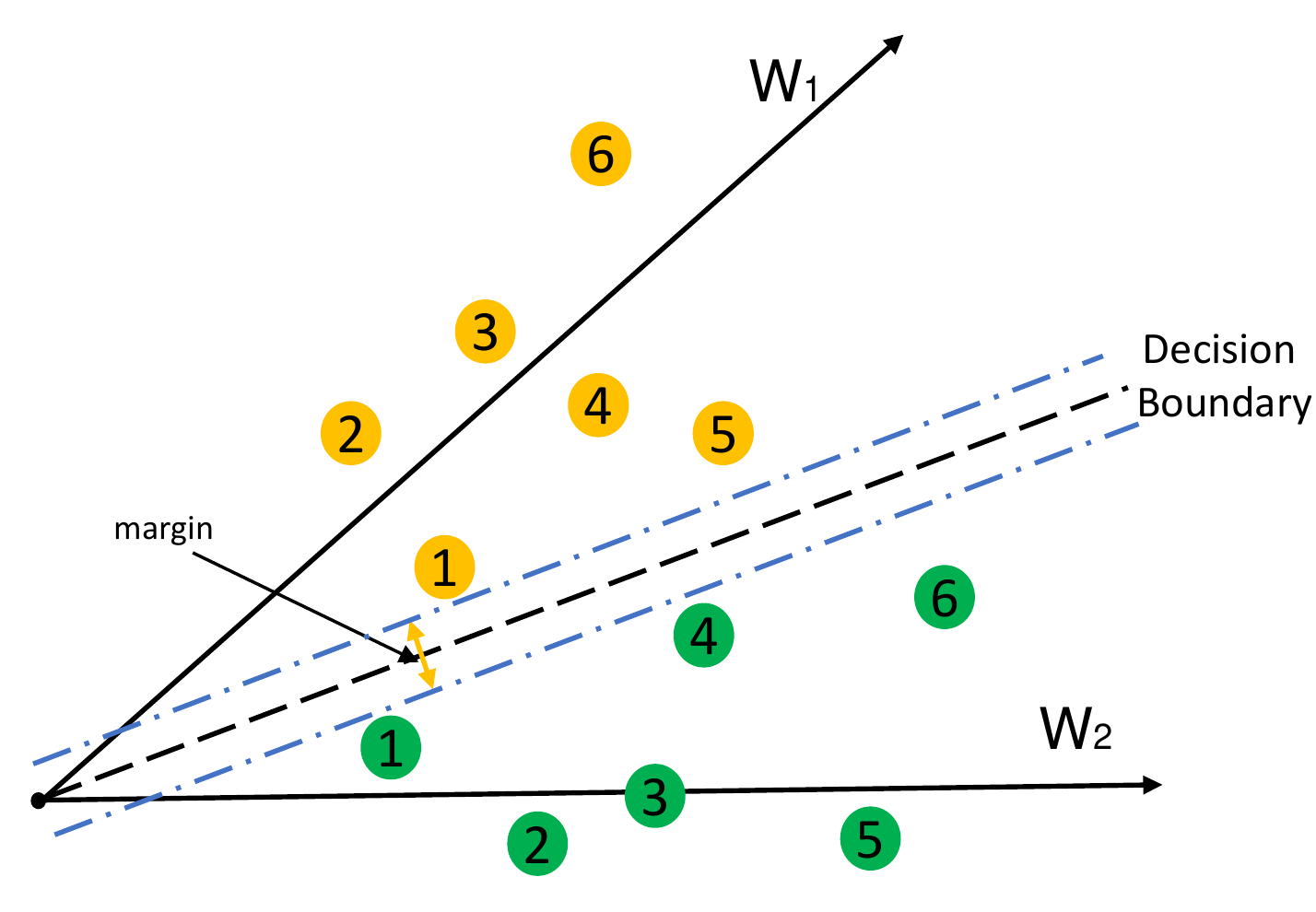}}
  \hspace{5mm}
}
\subfigure[]{
  \label{fig:fig1:margin2}
  {\includegraphics[width=4.0cm,bb=4 11 402 275]{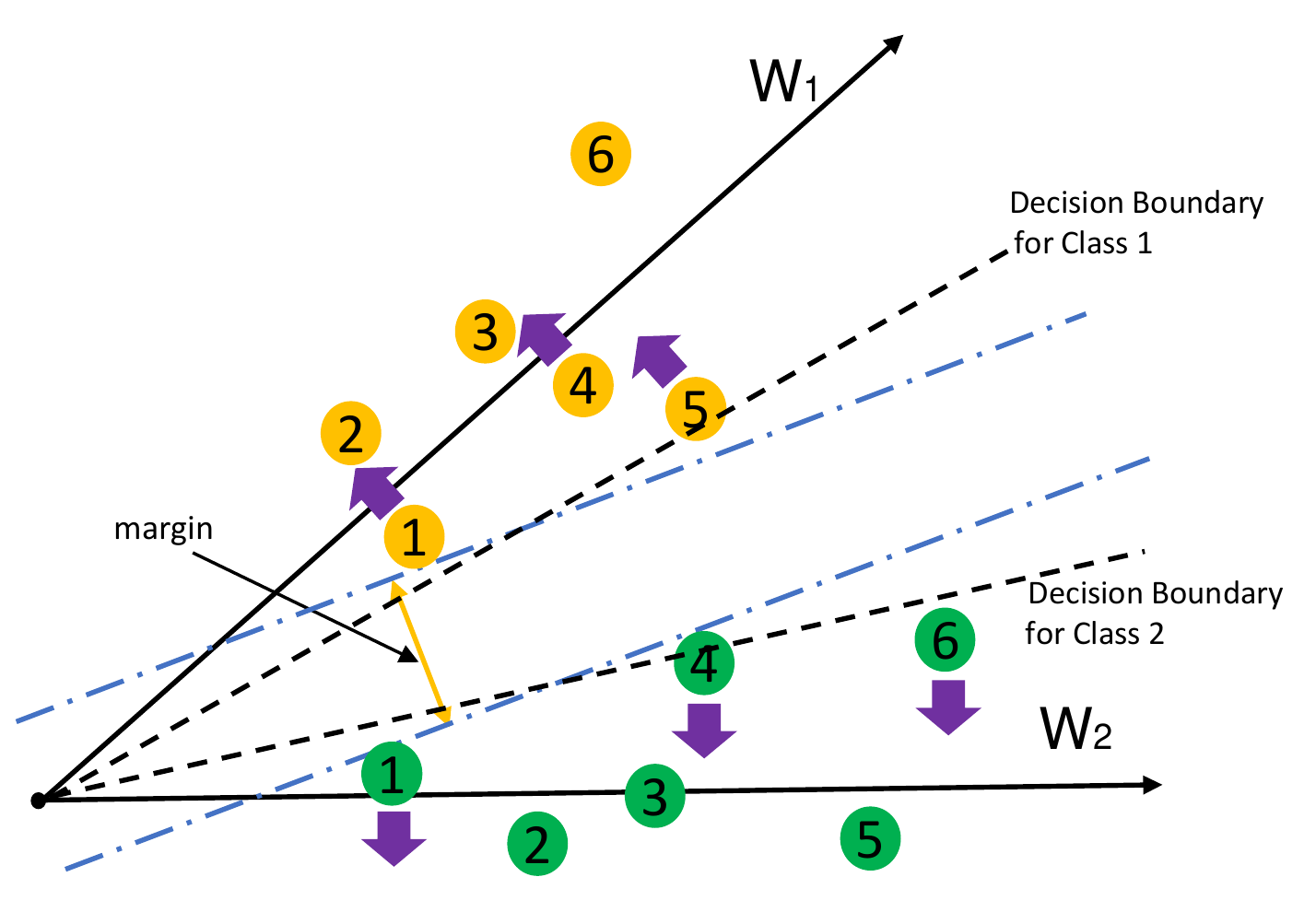}}
  \hspace{5mm}
}
\subfigure[]{
  \label{fig:fig1:margin3}
  {\includegraphics[width=4.0cm,bb=1 0 405 275]{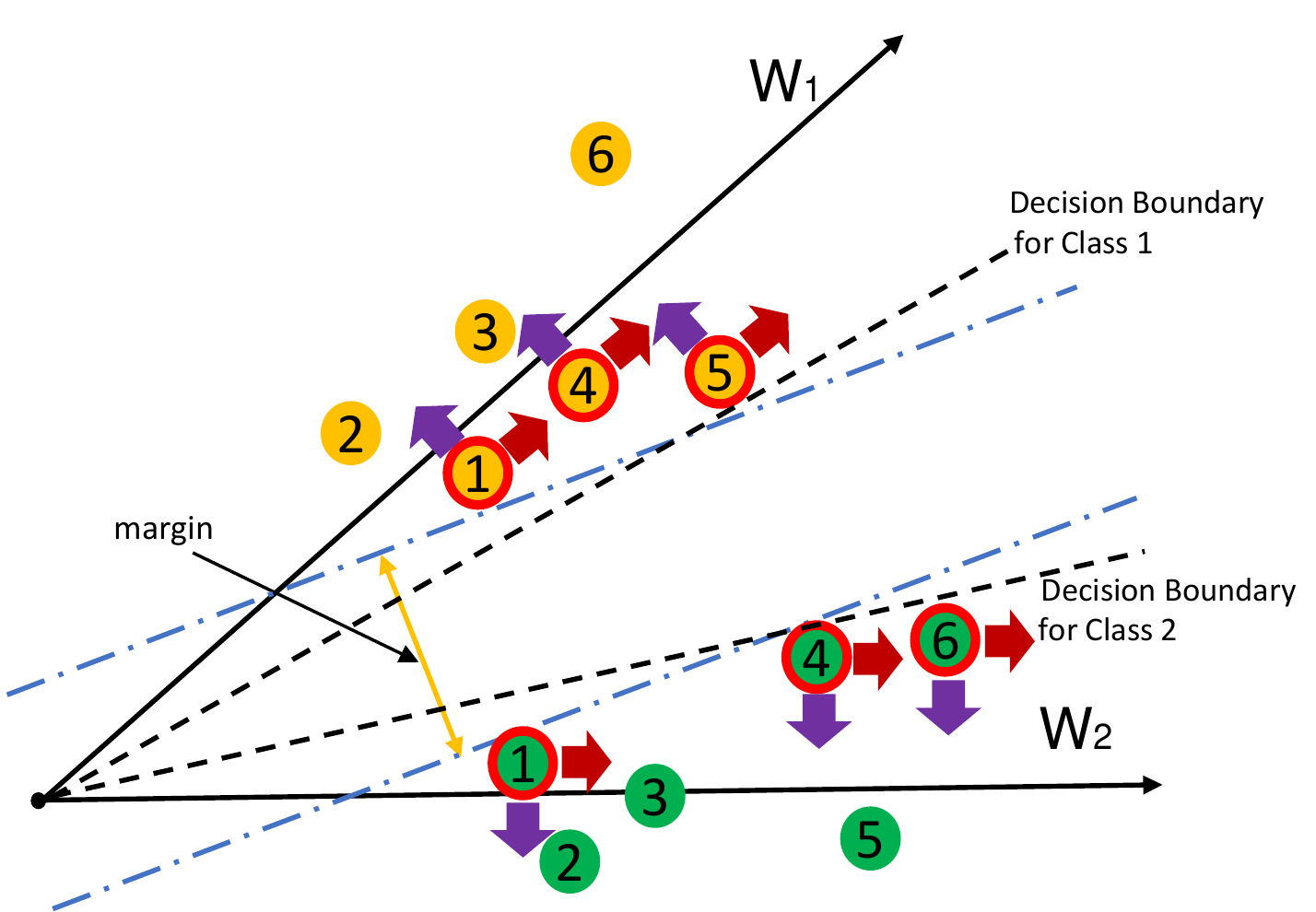}}
}
\caption{\footnotesize{Comparison of softmax loss with softmax loss + feature incay(i.e., Reciprocal Norm Loss) on CIFAR10, results are showed in subfigure (a)(b)(c). (a) Average $\emph{L}_{2}$-norm of feature vectors versus iterations on test set, (b) softmax loss versus iterations on test set, (c) top-1 accuracy versus iterations on test set. In (d)(e)(f), we schematically illustrate different losses using binary classification as example, where yellow points are samples of class1 and green points are samples of class2. The black dashed line represents the decision boundary between the two classes. The two blue dashed lines represent the hyperplanes that pass the points with minimal distances to the decision boundary.  The numbers 1-6 represent the increasing $\emph{L}_{2}$-norm of the points within each class. Their weight vectors within the softmax loss are $W_{1}$ and ${W_{2}}.$ (d) Feature embedding of softmax loss. (e) Feature embedding of large-margin softmax loss. The purple arrows represent the further constraints from large-margin softmax loss. Please refer ~\cite{liu2016large} for the detailed explanation about the two black dashed lines named Decision Boundary for Class1/Class2. (f) large-margin softmax loss + feature incay. The red arrows in (f) correspond to the loss constraints of feature incay. The margin between two classes increases from left to right by optimizing both the angles and $\emph{L}_{2}$-norm.}}
\label{fig:margin_}
\end{figure*}

Deep Neural Networks (DNNs) with softmax loss have achieved state-of-the-art performance on various multi-class classification tasks. In DNNs, both representation and classifier are learned simultaneously in a single network, where the final representation for a sample is the feature vector $\bold{f}$ outputted from the penultimate layer, while the last layer outputs scores $z_i = \bold{w}_i \cdot \bold{f}$ for each category $i$, where $\bold{w}_i$ is the weight vector for category $i$. To define softmax loss, the scores are further normalized into probabilities via softmax function, i.e., {\scriptsize{$p_i = \frac{e^{z_i}}{\sum_j e^{z_j}}$}}. A well-trained DNN should output highest probability for the correct label in both training and testing phases, which requires the score for the correct label is significantly larger than other labels. Since $z_i = \bold{w}_i \cdot \bold{f} = \|\bold{w}\| \|\bold{f}\|\cos(\theta)$, where $\theta$ is the angle between $\bold{w}$ and $\bold{f}$, the goal of significant larger score for the correct label than other labels can be achieved by tuning $\|\bold{w}\|$, $\|\bold{f}\|$ and $\theta$. While increasing the weight norm $\|\bold{w}\|$ is constrained by weight decay for regularization, then $\|\bold{f}\|$ and $\theta$ become the two main factors for optimization. Softmax loss optimizes both factors but leaves much room to improve of each factor.

To improve softmax loss, large-margin softmax~\cite{liu2016large} is proposed to further decrease the angle between the feature vector and the weight vector of the correct label by adding more rigid constraint on the angle. Motivated by the fact that $\emph{L}_{2}$-normalized feature vector often results in better distance metric for retrieval tasks, ~\cite{wang2017normface,liu2017sphereface} propose to use $\emph{L}_{2}$-normalized feature vector in softmax loss. Since feature vectors are normalized, feature norm has no affect on the score and only angle is optimized, which also results in much lower intra-class angular variability compared with softmax loss. All these methods explicitly or implicitly optimize the angle further but ignore the effect of feature norm. In this paper, feature incay is proposed to further optimize the factor of feature norm. In contrast to weight decay that favors weight vectors with small norm, feature incay favors feature vectors with large norm. From the computational perspective, larger feature norm results in larger score differences among categories, which can better separate the categories. From the perspective of pattern detection, larger feature norm encourages model to learn and detect more prominent patterns.

Feature incay is a general regularizer and can be added to softmax loss to emphasize feature norm. Whereas large-margin softmax emphasizes on the angle, feature incay can also be added to it to also emphasize on the feature norm. Specifically, feature incay is implemented by Reciprocal Norm Loss which minimizes the reciprocal of feature vectors' $\emph{L}_{2}$ norm.


Figure~\ref{fig:fig1:softmax_iter200},~\ref{fig:fig1:softmax_iter2000} and ~\ref{fig:fig1:softmax_iter20000} shows the empirical results by adding feature incay to softmax loss, where feature incay brings larger $\emph{L}_{2}$-norm, smaller training loss and higher accuracy on test set. The geometric interpretation of feature incay is illustrated in Figure ~\ref{fig:fig1:margin1}, ~\ref{fig:fig1:margin2} and ~\ref{fig:fig1:margin3}, large-margin softmax~\ref{fig:fig1:margin2} adds stronger constrain to align feature vectors to the weight vectors of their corresponding labels, by further adding feature incay~\ref{fig:fig1:margin2}, feature vectors are pushed away from the origin which results in better inter-class separability. Besides, Reciprocal Norm Loss is designed to increase the $\emph{L}_{2}$-norm adaptively according to the features' original $\emph{L}_{2}$-norm, which can also help reduce the intra-class variances as illustrated in Figure ~\ref{fig:property}. 

In summary, we analyze the effect of the features' $\emph{L}_{2}$-norm and prove (1) the benefit of the features' $\emph{L}_{2}$-norm, (2) why the softmax loss fails to optimize the $\emph{L}_{2}$-norm of the features that have been well classified and (3) how our method can ensure the inter-class separability and intra-class compactness. The proposed Reciprocal Norm Loss is verified on three widely used classification datasets, i.e., MNIST, CIFAR10 and CIFAR100 using various network structures. We also propose the General Reciprocal Norm Loss to replace the softmax loss with other loss functions, such as Center loss and large-margin softmax loss. By considering the feature incay, we have achieved state-of-the-art performance on CIFAR10 and CIFAR100.

\section{Related work}

\textbf{Large-margin Softmax Loss.} Liu et al.~\cite{liu2016large} proposed to improve softmax loss by incorporating a adjustable margin $m$ multiplying the angle between a feature vector and the corresponding weight vector. Compared with the softmax loss, the new loss function pays more attention to the angular decision margin between classes as illustrated in Figure~\ref{fig:fig1:margin2}. Large-margin softmax loss adding stronger constrain to the angular, while feature incay adding stronger constrain to feature norm which is orthogonal to large-margin softmax loss as illustrated in Figure ~\ref{fig:fig1:margin3}.

\textbf{Center Loss.} Wen et al.~\cite{wen2016discriminative} proposed the center loss to learn centers for deep features of each class and penalize the distances between the deep features and their corresponding class centers. The softmax loss tries to align features vectors close to the weight vectors based on the inner product similarity, while center loss pushes feature vectors towards their class centers according Euclidean distances. Combining softmax loss with center loss actually uses two sets of classifiers, where representation is learned based on both the inner product to weight vector and the Euclidean distance to class center. The added center loss helps minimize the intra-class distances also by influencing the $\emph{L}_{2}$-norm of feature vectors, namely, small $\emph{L}_{2}$-norm will be increased and large $\emph{L}_{2}$-norm will be decreased along the process of pushing feature vectors to class centers. Different from center loss, feature incay also increases the norm of feature vectors with large $\emph{L}_{2}$-norm instead of penalizing feature vectors with large norm as center loss.


\textbf{Weight/Feature Normalization.} Inspired by the fact that feature normalization before calculating the sample distance usually achieves better performance for retrieval tasks, Rajeev et al.~\cite{ranjan2017l2} proposed to use normalized feature vectors in softmax loss during training stage, thus the feature norm has no effect on softmax loss and angle is the main factor to be optimized. Congenerous cosine loss~\cite{liu2017learning}, NormFace~\cite{wang2017normface}, and cosine normalization~\cite{chunjie2017cosine} take a step further to normalize the weight vectors which replace inner product with cosine similarity within softmax loss, and only optimize the factor of angle. There normalization methods achieve much lower intra-class angular variability by emphasizing more on the angle during training, while ignore that feature norm is another useful factor worth to optimize.

\section{Our work}
%
%

\subsection{Formulation of softmax loss}

Let $\mathbb{X} = \{(x_i, y_i)\}_{i=1}^{N}$ be the training set contains $N$ samples, where $x_i$ is the raw input to the DNN, $y_{i} \in \{1,2,\cdots,K\}$ is the class label that supervises the output of the DNN. Denote $\bold{f}_i$ as the feature vector for $x_i$ learned by the DNN, $\{\bold{w}_j\}_{j=1}^{K}$ represent weight vectors for the $K$ categories. Then, softmax loss is defined as,

\begin{align}
\mathcal{L}_{\text{softmax}}=-\frac{1}{N}\sum_{i=1}^{N} \log\bigg( \frac {e^{\bold{w}_{y_i}^{T}\bold{f}_{i}+\bold{b}_{y_i}}} {\sum_{j=1}^{K}e^{\bold{w}_{j}^{T}\bold{f}_{i} + \bold{b}_{j}}} \bigg). \label{softmax_loss}
\end{align}
The bias term will be removed as it has been proven unnecessary in the recent works~\cite{liu2016large,wang2017normface}. Denote the angle between $\bold{w}_{j}$ and $\bold{f}_{i}$ as ${\theta}_{\bold{w}_{j}, \bold{f}_{i}}$, the inner product between $\bold{w}_{j}$ and $\bold{f}_{i}$ can be rewritten as
\begin{align}
\bold{w}_{j}^{T}\bold{f}_{i} = \|\bold{w}_{j} \|\| {\bold{f}_{i}} \| \cos({\theta}_{\bold{w}_{j}, \bold{f}_{i}}). \label{cos}
\end{align}
By combining the above two equations, we get
\begin{align}
\mathcal{L}_{\text{softmax}}=-\frac{1}{N}\sum_{i=1}^{N}\log\bigg( \frac {e^{\| \bold{w}_{y_i} \| \| \bold{f}_{i} \| \cos({\theta}_{\bold{w}_{y_i}, \bold{f}_{i}})}} {\sum_{j=1}^{K}e^{\| \bold{w}_{j} \| \| \bold{f}_{i} \| \cos({\theta}_{\bold{w}_{j}, \bold{f}_{i}})}} \bigg) \label{softmax_loss_cos}
\end{align}

\subsection{Properties of feature norm $\| \bold{f}_{i} \|$}

Here are some properties that state why feature norm matters and why there is still improvement room for softmax loss. Before introducing these properties, we first introduce a \textbf{Lemma} that was proposed by~\cite{ranjan2017l2}:

\textbf{Lemma} \emph{When the number of classes $K$ is smaller than twice the feature dimension $D$, we can distribute the classes on a hypersphere of dimenstion $D$ such that any two class weight vectors are at least $90^{\circ}$ apart.}

The first property is similar to the proposition proved by~\cite{wang2017normface}, which states the relation between feature norm and softmax loss.

\textbf{Property 1} \emph{By fixing weight vectors and directions of the feature vectors, softmax loss is a monotonically decreasing function with the increasing of the features' $\emph{L}_{2}$-norm as long as the features are correctly classified.}

\textbf{\emph{Proof.}} Let $\mathcal{L}_{\text{softmax}}(\bold{f}_i)$ represent the loss of the $i$-th sample, $i = 1, \cdots, N$. Specifically,
\begin{align}
\mathcal{L}_{\text{softmax}}(\bold{f}_i)&=-\log\bigg( \frac {e^{\| \bold{w}_{y_i} \| \| \bold{f}_{i} \| \cos({\theta}_{\bold{w}_{y_i}, \bold{f}_{i}})}} {\sum_{j=1}^{K}e^{\| \bold{w}_{j} \| \| \bold{f}_{i} \| \cos({\theta}_{\bold{w}_{j}, \bold{f}_{i}})}} \bigg)  \notag \\
&= -\log\bigg( \frac {1} {\sum_{j=1}^{K}e^{\| \bold{w}_{j} \|\| \bold{f}_{i} \| \cos({\theta}_{\bold{w}_{j}, \bold{f}_{i}}) - \| \bold{w}_{y_i} \|\| \bold{f}_{i} \| \cos({\theta}_{\bold{w}_{y_i}, \bold{f}_{i}})}} \bigg)
\end{align}
Recall that when $\bold{f}_{i}$ is correctly classified, we have $\bold{w}_{y_i}^{T}\bold{f}_{i} > \bold{w}_{j}^{T}\bold{f}_{i}$ for any $j \neq y_{i}$, and $\| \bold{w}_{j} \|\| \bold{f}_{i} \| \cos({\theta}_{\bold{w}_{j}, \bold{f}_{i}}) - \| \bold{w}_{y_i} \|\| \bold{f}_{i} \| \cos({\theta}_{\bold{w}_{y_i}, \bold{f}_{i}}) \le 0$ always holds. Then, for any $t > 0$, we have
\begin{align}
\mathcal{L}_{\text{softmax}}((1 + t)\bold{f}_i) < \mathcal{L}_{\text{softmax}}(\bold{f}_i),
\end{align}
which means increasing the norm of correctly classified samples can decrease the softmax loss. To consider all samples including incorrectly classified ones, we set $t_i > 0$ if $i$ is correctly classified and $t_i=0$ otherwise, then we have
\begin{align}
\sum_{i=1}^{N} \mathcal{L}_{\text{softmax}}((1 + t_i)\bold{f}_i) \leq \sum_{i=1}^{N} \mathcal{L}_{\text{softmax}}(\bold{f}_i)
\end{align}
So feature norm is an important factor to achieve smaller softmax loss together with the angle.
$\hfill{}\Box$

Though the first property states that minimizing softmax loss encourages feature vectors towards large norm, there is still improvement room to achieve larger norm.

\textbf{Property 2} \emph{For any feature vector $\bold{f}_{i}$, when the probability $P_{y_i}$ of the right category $y_i$ outputted by softmax function for $\bold{f}_{i}$ is close to one, the gradient of $\bold{f}_{i}$ computed based on softmax loss will be close to zero, which leads that the intra-class compactness and inter-class separability could not be further improved.}

\textbf{\emph{Proof.}} According to definition of softmax loss in Eq.\eqref{softmax_loss_cos}, the gradient of feature vector $\bold{f}_{i}$ is:
\begin{align}
\frac{\partial \mathcal{L}_{\text{softmax}}}{\partial \bold{f}_{i}} =  \frac{1}{N} (- \bold{w}_{y_{i}} + \sum_{j=1}^{K} P_{j} \bold{w}_{j} )
\end{align}
where the $P_{j} = \frac{e^{\bold{w}_{j}^{T}\bold{f}_{i}}}{\sum_{j=1}^{K}e^{\bold{w}_{j}^{T}\bold{f}_{i}}}$.
When $P_{y_i} \to 1$ and $P_{j} \to 0  (\forall j \neq y_{i})$, $\frac{\partial \mathcal{L}_{\text{softmax}}}{\partial \bold{f}_{i}} \to 0$. That is after a training sample is confidently classified correctly, it will have no contribution to its own representation learning.
For example, suppose $\| \bold{w}_{j} \| = 1$, ${\theta}_{\bold{w}_{y_i}, \bold{f}_{i}}  = 0$ and
${\theta}_{\bold{w}_{j}, \bold{f}_{i}} \ge 90^{\circ} (\forall j \neq y_{i})$ according the \textbf{Lemma}, then $\| \bold{w}_{y_i} \| \| \bold{f}_{i} \| \cos({\theta}_{\bold{w}_{y_i}, \bold{f}_{i}}) = \| \bold{f}_{i} \|$, $\| \bold{w}_{j} \| \| \bold{f}_{i} \| \cos({\theta}_{\bold{w}_{j}, \bold{f}_{i}}) < 0, \forall j \neq y_{i}$. Putting together, we have
$P_{y_i} \geq \frac{e^{\| \bold{f}_{i}\|}}{e^{\| \bold{f}_{i}\|} + K-1}$, for a modest number of categories say $K=10$, $P_{y_i} > 0.999$ when $\| \bold{f}_{i} \| = 10$. More categories can optimize to larger feature norm before softmax is saturation.

$\hfill{}\Box$

This property says feature norm can still be elongated even the softmax loss is 0. The Reciprocal Norm Loss discussed in next section is to further elongate the feature norm explictly.

%
%

\subsection{Reciprocal Norm Loss}
Reciprocal Norm Loss adds a term of feature incay to penalize feature vectors with small norm and formally defined as
\begin{align}
\quad \mathcal{L} = \underbrace{-\frac{1}{N}\sum_{i=1}^{N}\log\bigg( \frac {e^{\bold{w}_{y_i}^{T}\bold{f}_{i}}} {\sum_{j=1}^{K}e^{\bold{w}_{j}^{T}\bold{f}_{i}}} \bigg)}_{\text{softmax loss}} +  \underbrace{\mu \sum_{k=1}^{K} \| \bold{w}_{k} \|^{2}}_{\text{weight decay}} + \underbrace{\lambda \frac{1}{N}\sum_{i=1}^{N} \frac{h(i)}{\| \bold{f}_i \|^{2} +  \epsilon}}_{\text{feature incay}}
\label{rn_loss}
\end{align}

where the $h(i)$ is an indicator function, and $h(i) = 1$ when the $\bold{f}_i$ is correctly classified otherwise $h(i) = 0$, $\epsilon$ is a small positive value to prevent dividing by value close to zero, $\lambda$ is a hyperparameter used to control the influence of feature incay. The whole loss function consists of three parts: softmax loss, weight decay and feature incay, and together they favor weights that achieve small softmax loss, small weight norm and large feature norm. In the early stage of training, most $h(i)$ are equal to 0, and softmax loss and weight decay are active in training, and feature incay is gradually adding in along with training to reinforce the feature vectors that result in correct classification. Large feature norm brings large inter-class separability under the constrain of small weight norm, otherwise large feature norm can be trivially achieved by increasing the weight norm. The simple reciprocal form of feature norm has another nice property that moves feature vectors with small norm fast while moves feature vectors with large norm slow, which results intra-class compactness. Specifically, denote feature incay of $\bold{f}_{i}$ as $\mathcal{F}(\bold{f}_i) = \frac{h(i)}{\| \bold{f}_{i}\|^{2} + \epsilon}$, when $h(i) = 1$, the gradient is $\frac{\partial \mathcal{F}}{\partial \bold{f}_i} = -\frac{2\bold{f}_i}{(\| \bold{f}_i \|^{2} + \epsilon)^2}$. For any two feature vectors $\bold{f}_{p}$ and $\bold{f}_{q}$ satisfying $\| \bold{f}_{q} \| \hspace{1mm}> \| \bold{f}_{p} \|$, we always have $\|\frac{\partial \mathcal{F}}{\partial \bold{f}_{p}}\| \hspace{1mm}> \|\frac{\partial \mathcal{F}}{\partial \bold{f}_{q}}\|$. That is feature vectors with small norm increase fast along their original directions while feature vectors with large norm increase slowly along their original directions. 


\begin{figure*}
\hspace{2mm}
\centering
\includegraphics[width=9.0cm,bb=54 0 701 144]{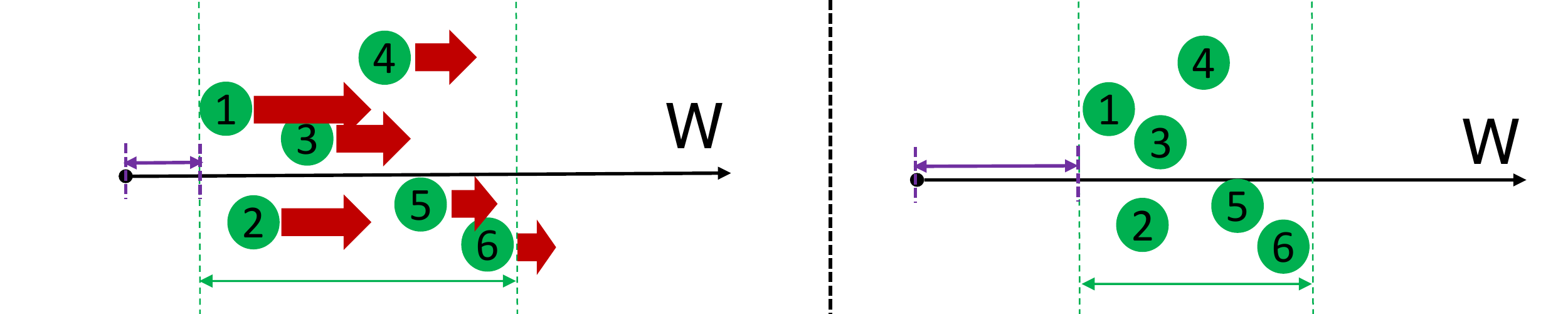}
\caption{\footnotesize{The original data distribution is on the left of the black dashed line and the data distribution updated according to the Reciprocal Norm Loss is on the right. The numbers 1-6 represent the points are of increasing feature norm. The black point represents the original point.  The lengthes of the green bidirectional arrows represent the maximal distance in the direction of the weight vectors within all the points of one class. The purple bidirectional arrow means the minimal distance to origin, which is equal to the minimal feature norm. The red arrows represent the gradients update along the directions of the weight vectors computed with the Reciprocal Norm Loss, while the length represents the size of the gradient.}}
\label{fig:rn}
\end{figure*}


\subsection{Geometric Interpretation of Reciprocal Norm Loss}
\begin{wrapfigure}{l}{5cm}
\centering
\includegraphics[width=4.0cm,bb=45 27 402 396]{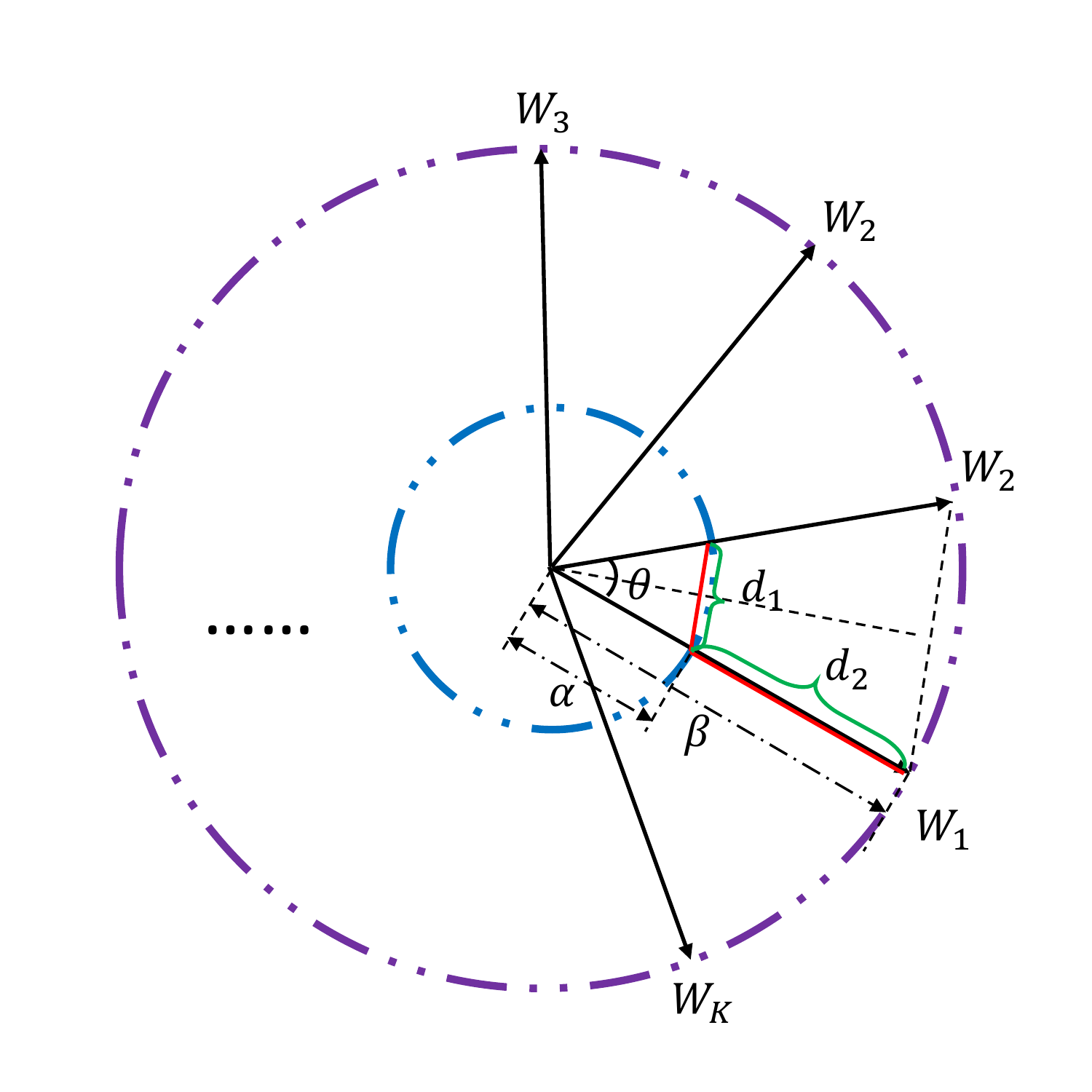}
\caption{\footnotesize{Illustration in 2-dimensional space.}}
\label{fig:property}
\end{wrapfigure}
\textbf{Property 3} Suppose (a)\hspace{1mm}the angle between any feature vector $\bold{f}_{i}$ and its corresponding weight vector $\bold{w}_{y_i}$ is zero, (b)\hspace{1mm}the angles between any two neighbor weight vectors of different classes are $\theta$, we have (1) the minimal inter-class distance is $2\alpha \sin(\frac{\theta}{2})$, where $r$ is lower bound of feature norm, (2) when $K < 2D$, the upper bound of feature norm is in the range of $[(1+\sqrt{2})\alpha, 3\alpha]$ to ensure the maximal intra-class distance is smaller than the minimal inter-class distance.


\textbf{\emph{Proof.}} Figure~\ref{fig:property} shows the 2-dimensional case satisfying the conditions(a)(b), where black arrows named with $W_{i}$ represent weight vectors for each class. As we have assumed all feature vectors are lying on the directions of their corresponding $W_{i}$, and blue circle and purple circle denote the lower bound and upper bound of feature norm respectively. Thus the maximal intra-class distance is $d_{2} = \beta - \alpha$ and the minimal inter-class distance is $d_{1} = 2\alpha \sin(\frac{\theta}{2})$. To ensure minimal inter-class distance is larger than intra-class distance, i.e., $d_1 = 2\alpha \sin(\frac{\theta}{2}) > d_2 = \beta - \alpha$, which requires $\beta < 2\alpha \sin(\frac{\theta}{2}) + \alpha$.

When $K < 2D$, according to the \textbf{Lemma}, we can ensure that $\theta \ge 90^{\circ}$. Besides, the angle between any two vectors is smaller than $180^{\circ}$. Then $\frac{\theta}{2} \in [45^{\circ}, 90^{\circ}]$ and $\sin(\frac{\theta}{2})$ is a monotonously increasing function within the range $[45^{\circ}, 90^{\circ}]$. Based on $\sin(\frac{\theta}{2})\in [\frac{\sqrt{2}}{2}, 1]$, the upper bound of $\emph{L}_{2}$-norm of feature vectors is in the range $[(1+\sqrt{2})\alpha, 3\alpha]$.

$\hfill{}\Box$

According to the \textbf{Property}, the feature norm does not need to increase endlessly, which can also be achieved by feature incay since feature vectors with large norm have negligible effect on the model. we can estimate an upper bound according to the original features' $\emph{L}_{2}$-norm. In our experiments, we will choose the average $\emph{L}_{2}$-norm as the lower bound to avoid the outlier.

In summary, besides information in training examples and complexity of the neural network~\cite{krogh1992simple}, feature norm can also influence the generalization ability.

\section{Experiments}
In this section, we verify the effectiveness of feature incay through empirical experiments.

\subsection{Experimental Settings}
We evaluate feature incay on three datasets, i.e., MNIST, CIFAR10, and CIFAR100. MNIST consists of 60,000 training images and 10,000 test images from 10 handwritten digits, both CIFAR10 and CIFAR100 contain 50,000 training images and 10,000 test images from 10 object categories and 100 object categories respectively. Images are horizontally flipped randomly and subtracted by mean image. Neural networks architectures used in~\cite{liu2016large} are adopt with minor modification as detailed in Table ~\ref{table:cnn_architecture}, and implemented based on Caffe framework~\cite{jia2014caffe}. The weight $\mu$ for weight decay is 0.0005, the weight $\lambda$ is set with different values in different experiments, e.g., 1.0, 0.1 or 0.01. The momentum is 0.9, and the learning rate starts from 0.1 and divided by a factor of 10 three times when the training error stops decreasing.


\begin{table*}[!htb]
\footnotesize
\centering
\caption{\footnotesize{The CNN architectures used for MNIST/CIFAR10/CIFAR100}. The count of the Conv1.x, Conv2.x and Conv3.x closely follows the settings in ~\cite{liu2016large}. All the pooling layers are with window size $2\times2$ and stride of 2.}
\begin{tabular}{l|c|c|c|c}
\hline
Layer & MNIST-2D & MNIST & CIFAR10 & CIFAR100 \\ \hline
Conv0.x & $\lbrack 5\times5,20\rbrack \times 1$ & $\lbrack 3\times3,64\rbrack  \times 1$  & $\lbrack 3\times3,64\rbrack  \times 1$ & $\lbrack 3\times3,128\rbrack  \times 1$\\
\hline
Conv1.x & $\lbrack 5\times5,50\rbrack  \times 1$ & $\lbrack 3\times3,64\rbrack  \times 3$ & $\lbrack 3\times3,64\rbrack  \times 4$ & $\lbrack 3\times3,128\rbrack \times 4$ \\
\hline
Conv2.x & N/A & $\lbrack 3\times3,64\rbrack \times 3$ & $\lbrack 3\times3,128\rbrack \times 4$ & $\lbrack 3\times3,256\rbrack \times 4$\\
\hline
Conv3.x & N/A & $\lbrack 3\times3,64\rbrack \times 3$ & $\lbrack 3\times3,256\rbrack \times 4$ & $\lbrack 3\times3,512\rbrack \times 4$\\
\hline
Fully Connected & 2 & 256 & 512 & 512 \\
\hline
\end{tabular}
\label{table:cnn_architecture}
\end{table*}

\begin{wraptable}{l}{7cm}
\footnotesize
\caption{\footnotesize{Results on MNIST.}}
\begin{tabular}{|l|c|c|c}
\hline
Method & Accuracy & Average $\emph{L}_{2}$-norm \\ \hline \hline
Softmax & 88.62 & 316.65\\
RN + Softmax & 89.44 & 488.86\\ \hline \hline
L-Softmax & 89.91 & 329.92\\
RN + L-Softmax &  92.06 & 552.54\\ \hline \hline
Center Loss & 89.01 & 192.15\\
RN + Center Loss &  91.32 & 273.78\\
\hline
\end{tabular}
\label{table:baseline_exp}
\end{wraptable}

\subsection{Quick Experiments}
To quickly evaluate the effectiveness of feature incay, we experiment on MNIST using a very simple neural network with only two convolutional layers as listed in Table~\ref{table:cnn_architecture}. Due the simplicity of the model and the task, the training error can be converged in only 10,000 iterations.
Table~\ref{table:baseline_exp} summarizes the top-1 accuracy and average $\emph{L}_{2}$-norm on test set by using Softmax, large-margin Softmax (L-Softmax), Softmax with center loss (Center Loss), and with feature incay (RN+). L-Softmax, Center Loss and RN+Softmax improve Softmax, and RN+L-Softmax and RN+Center Loss improve L-Softmax and Center Loss further. With explicitly penalizing feature vectors with small norm, feature incay significantly increases the average feature norm as expected.


\subsection{Comparison Experiments}
To achieve state-of-the-art performance on the three datasets, we use deep neural networks as specified in last three columns. Feature incay is added to Sofmax and L-Softmax to compare with state-of-the-art approaches, and the reproduced results by Softmax and L-Softmax following~\cite{liu2016large} are slightly better than the referred numbers in general. Table \ref{table:sota_exp} lists the error rates of compared approaches and our method. In can be concluded that feature incay can consistently improve over Softmax, and improves L-Softmax on CIFAR10 and CIFAR100, while slightly worse than our reproduced L-Softmax on MNIST, we hypothesis that the performance on MNIST is already saturate and difficult to improve further.

\begin{table*}
\centering
\caption{\footnotesize{Error rate (\%) on MNIST/CIFAR10/CIFAR100}}
\begin{tabular}{l|c|c|c}
\hline
Method & MNIST & CIFAR10 & CIFAR100\\ \hline \hline
CNN~\cite{jarrett2009best} & 0.53 & N/A & N/A \\
DropConnect~\cite{wan2013regularization} & 0.57 & 9.41 & N/A\\
FitNet~\cite{romero2014fitnets} & 0.51 & N/A & 35.04\\
NiN~\cite{lin2013network} & 0.47 & 10.47 & 35.68\\
Maxout~\cite{goodfellow2013maxout} & 0.45 & 11.68 & 38.57\\
DSN~\cite{lee2015deeply} & 0.39 & 9.69 & 34.57\\
R-CNN~\cite{liang2015recurrent} & 0.31 & 8.69 & 31.75\\
GenPool~\cite{lee2016generalizing} & 0.31 & 7.62 & 32.37\\
Hinge Loss~\cite{liu2016large} & 0.47 & 9.91 & 33.10\\
Softmax~\cite{liu2016large} & 0.40 & 9.05 & 32.74\\
L-Softmax~\cite{liu2016large} & 0.31 & 7.58 & 29.53\\ \hline \hline
Softmax & 0.35 & 8.45 & 32.36\\
RN + Softmax & 0.31 & 7.96 & 31.76\\\hline \hline
L-Softmax & \textbf{0.25} & 7.55 &  29.95\\
RN + L-Softmax & 0.29 & \textbf{7.32} & \textbf{29.18}\\
\hline
\end{tabular}
\label{table:sota_exp}
\end{table*}

\begin{figure*}[htb]
\hspace{-10mm}
\subfigure[]{
  \label{fig:exp:cifar10_acc}
  {\includegraphics[width=7.5cm,bb=25 13 537 405]{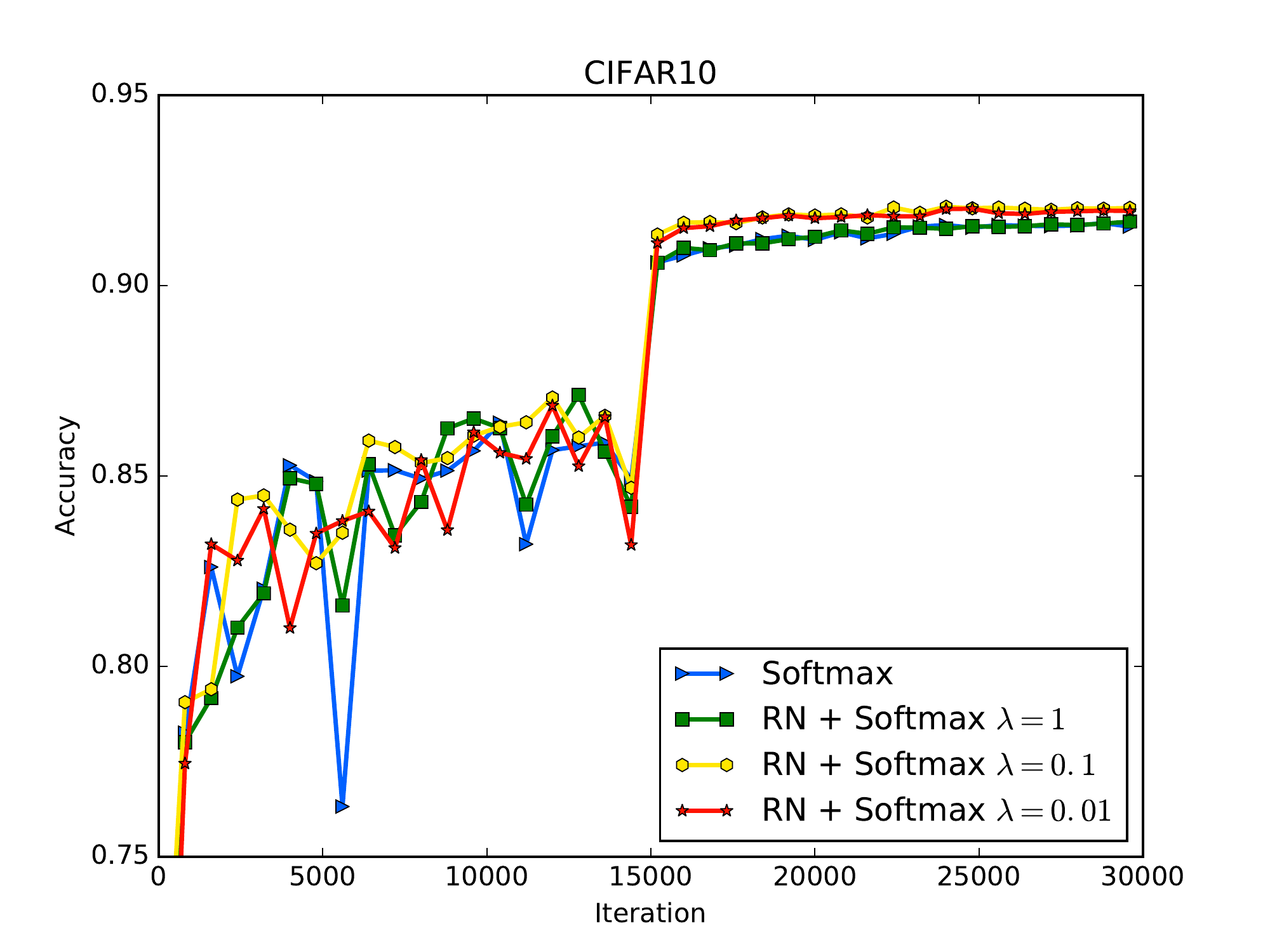}}
  \hspace{2mm}a
}
\subfigure[]{
  \label{fig:exp:cifar10_norm}
  {\includegraphics[width=7.5cm,bb=29 14 544 415]{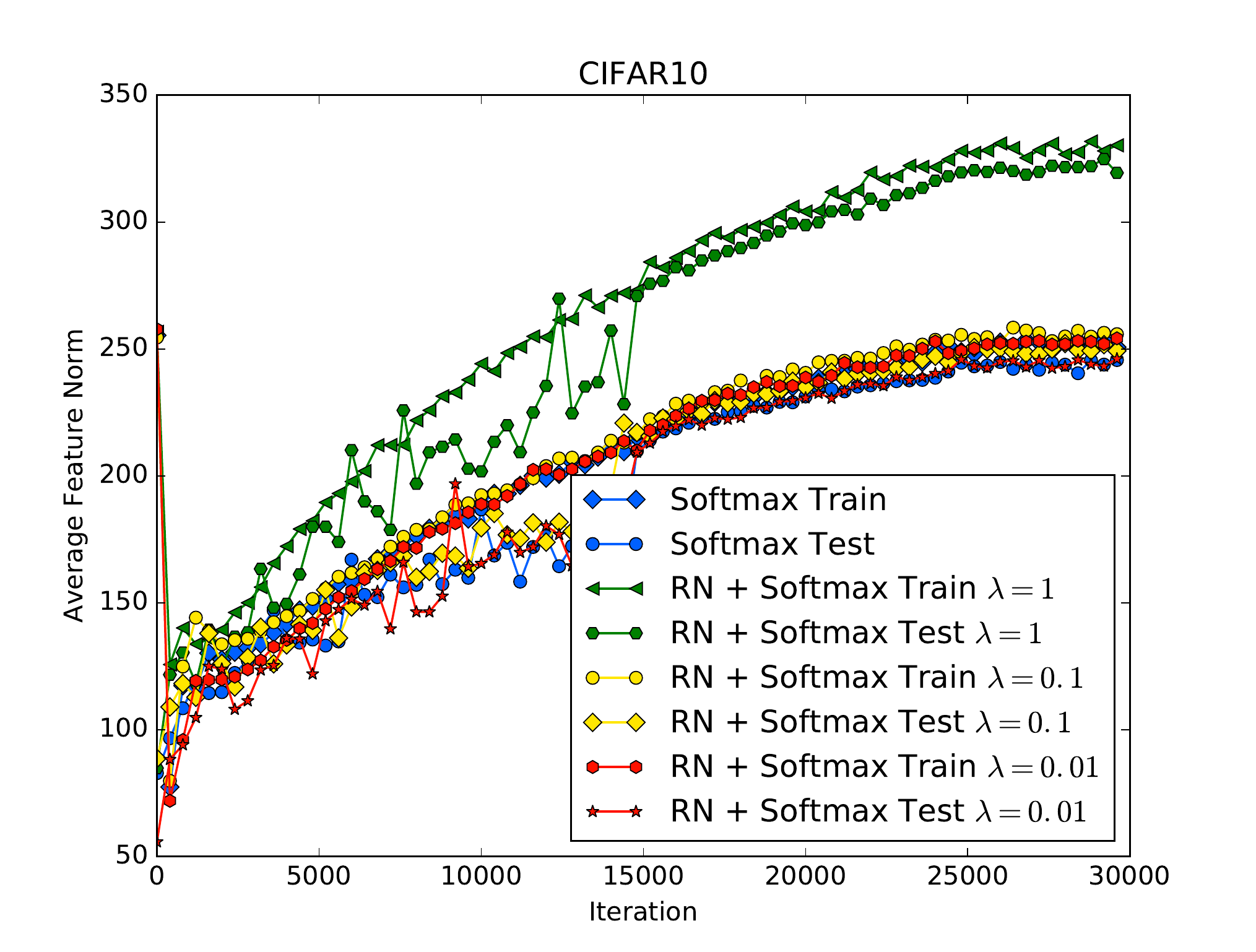}}
}
\subfigure[]{
  \hspace{-10mm}
  \label{fig:exp:cifar10_loss}
  {\includegraphics[width=7.5cm,bb=37 18 575 448]{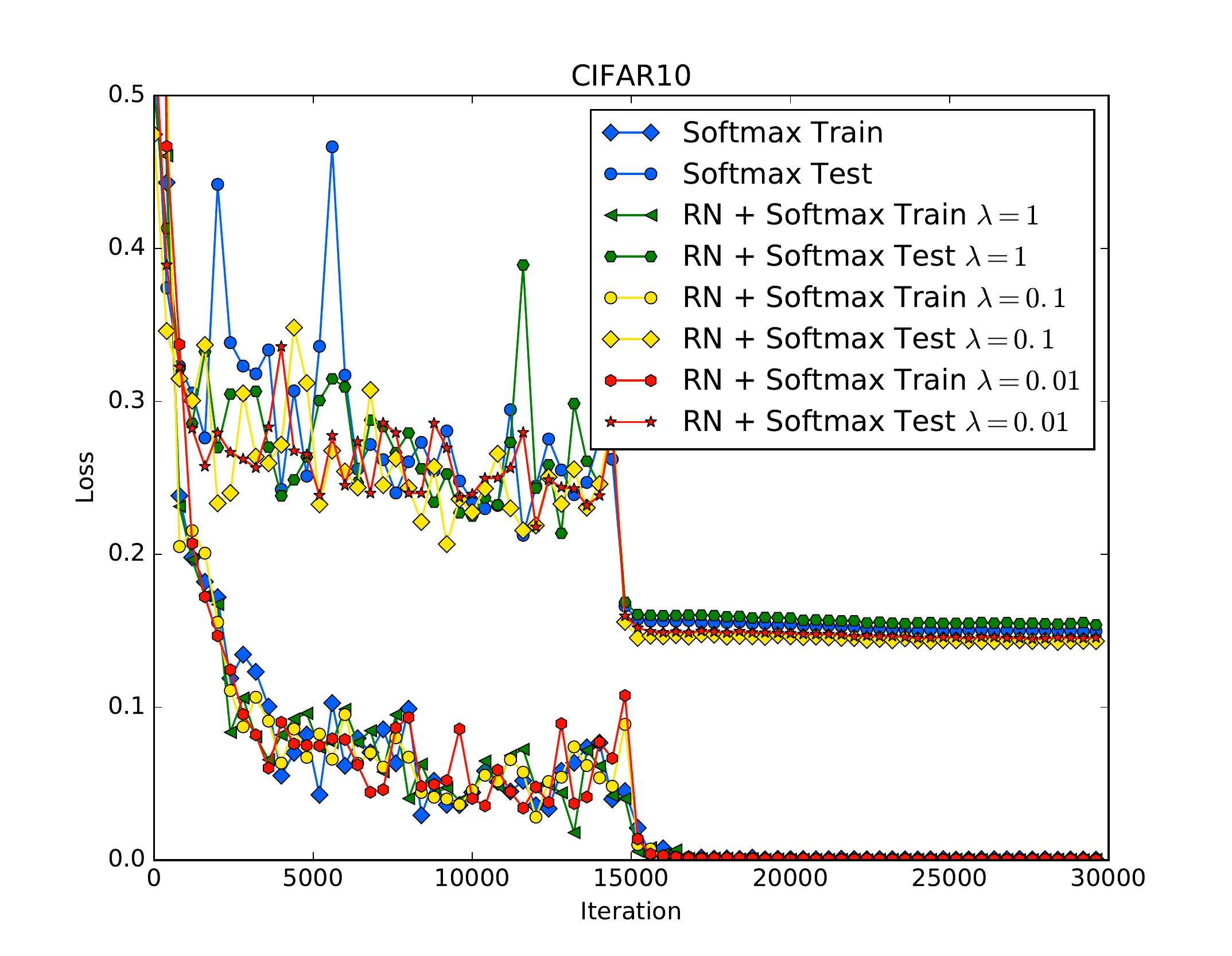}}
  \hspace{2mm}
}
\subfigure[]{
  \label{fig:exp:cifar100_acc}
  {\includegraphics[width=7.5cm,bb=25 13 537 405]{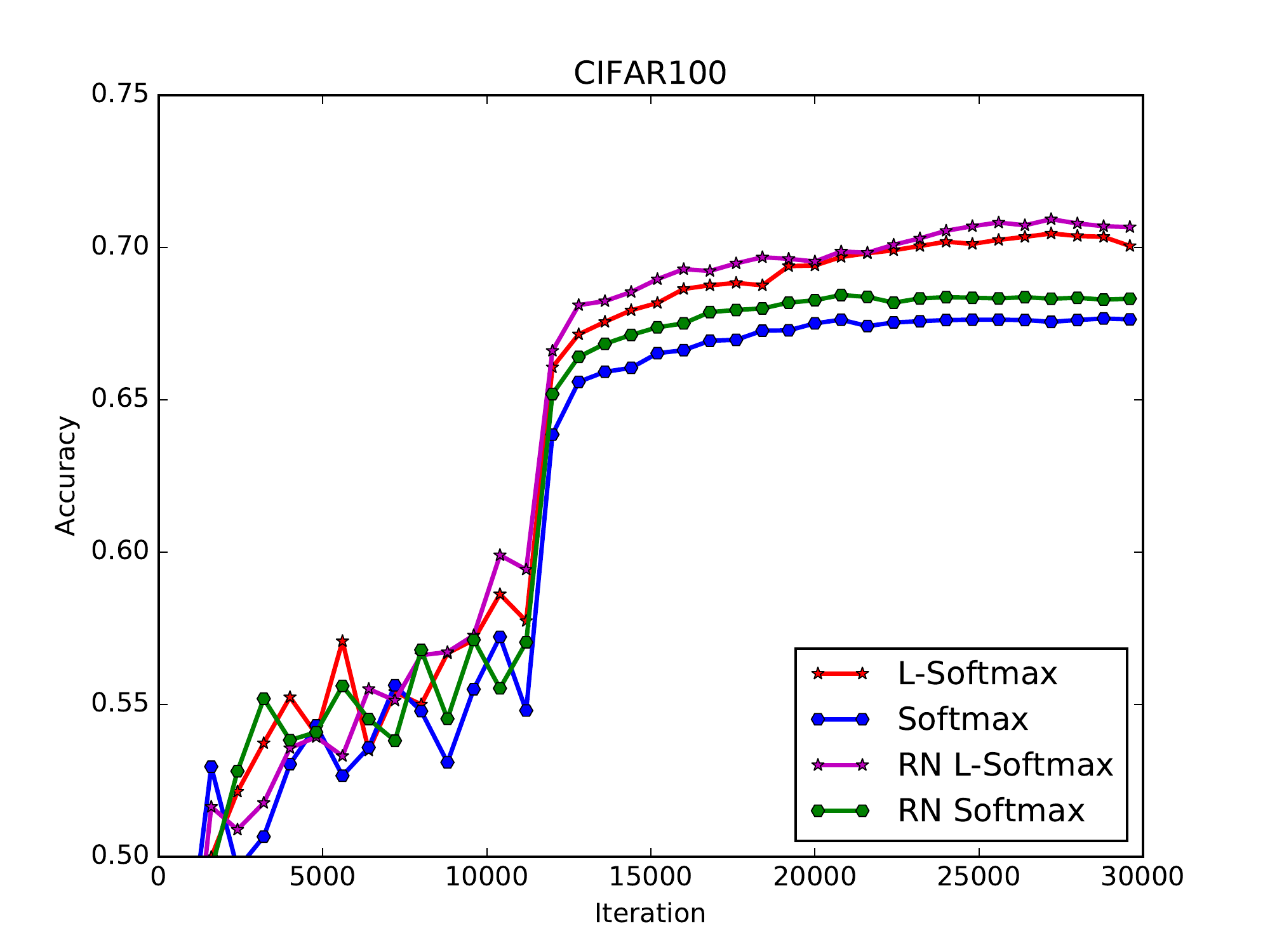}}
}
\caption{\footnotesize{(a) Accuracy versus iterations with different choices of $\lambda$ value on the test set of CIFAR10. The RN Softmax achieves 92.04\% when $\lambda=0.1$ (b) The training/testing sets' $\emph{L}_{2}$ norm versus iterations with different choices of the $\lambda$ value on CIFAR10. The RN Softmax with different $\lambda$ all achieve larger $\emph{L}_{2}$-norm. (c) The training/testing sets' loss vs. iterations with different choices of the $\lambda$ value on CIFAR10. The RN Softmax achieves notable smaller loss value 0.1432 than the Softmax with 0.1498. (d) Accuracy vs. iterations with Softmax/RN Softmax/L-Softmax/RN L-Softmax on CIFAR100.}}
\end{figure*}

Figure~\ref{fig:exp:cifar10_acc}, \ref{fig:exp:cifar10_norm} and \ref{fig:exp:cifar10_loss} show the accuracy, average feature norm and softmax loss during training by using $\lambda=1, 0.1, 0.01$ respectively. Feature incay achieves better or comparable accuracy compared with softmax loss under a wide range of $\lambda$, and results in larger feature norm on both training and test set. All methods achieve close to zero softmax loss on training set, while feature incay ensures lower softmax loss on test set. Figure \ref{fig:exp:cifar100_acc} shows the accuracy versus iteration number on CIFAR100, which is similar to the results on CIFAR10.

\subsection{Effects of $\lambda$}
\begin{table}
\centering
\caption{{Accuracy(\%) Comparison of different $\lambda$}}
\begin{tabular}{l|c|c|c|c}
\hline
Method & $\lambda=0$ & $\lambda=1$ & $\lambda=0.1$ & $\lambda=0.01$ \\ \hline \hline
RN + Softmax& 91.55 & 91.68 & \textbf{92.04} & 91.96 \\
RN + L-Softmax & 92.45 & 92.36 & 92.38 & \textbf{92.68} \\
\hline
\end{tabular}
\label{table:lambda_exp}
\end{table}
We conduct experiments on CIFAR10 to investigate how to choose the hyperparamter $\lambda$. Results are show in Table \ref{table:lambda_exp}. Softmax loss with feature incay achieves consistent improvement for all the different choices of $\lambda$. For adding feature incay to L-Softmax, only small $\lambda$ could improve the performance. This is due to L-Softmax emphasize more on the angle and often results in smaller feature norm than softmax loss, and the feature incay will be much larger, to balance the L-Softmax loss and feature incay for training, $\lambda$ should be set to a relatively small number.

\section{Conclusions and future work}
In this paper, we propose a novel Reciprocal Norm Loss as a new kind of regularizer named feature incay considering feature vectors are also tunable in DNNs. Feature incay reinforces feature vectors that result in correct classification, and elongate feature vectors with small norm fast than feature vectors with large norm, thus can decrease intra-class distance and increase inter-class distance at the same time. We also give the theoretical proof of why the feature norm matters, why there is still improvement room for the widely used softmax loss. Extensive experiments on MNIST, CIFAR10 and CIFAR100 verify the effectiveness of our method. In the future work, we plan to investigate other substitute feature incay and mine the inner relations between the feature incay and the weight decay.
 

{\small
\bibliographystyle{plain}
\bibliography{egbib}
}

\end{document}